\documentclass[times, review, 10pt]{elsarticle}

\usepackage{hyperref}
\usepackage{subfigure}
\usepackage{multirow}
\usepackage{color}
\journal{Journal of \LaTeX\ Templates}
\usepackage{framed,multirow}
\usepackage{amsmath,amsfonts}
\usepackage{makecell}
\usepackage{balance}
\usepackage{url}
\usepackage{algorithm}
\usepackage{algorithmic}

\begin{document}

\begin{frontmatter}

\title{Adaptive Confidence-Wise Loss for Improved Lens Structure Segmentation in AS-OCT}

\author[1]{Zunjie~Xiao}
\author[1]{Xiao~Wu\corref{cor1}}

\author[1]{Tianhang Liu}
\author[1]{Lingxi~Hu}

\author[1]{Yinling~Zhang}
\author[1]{Xiaoqing~Zhang}
\author[1,2,3]{Risa~Higashita}
\author[1,3,4]{Jiang Liu\corref{cor2}}
\cortext[cor1]{Contribute equally.}
\cortext[cor2]{Corresponding author.\\ 
  E-mail address:liuj@sustech.edu.cn (J.Liu).}


\address[1]{Research Institute of Trustworthy Autonomous Systems and Department of Computer Science and Engineering, Southern University of Science and Technology, Shenzhen 518055, China}

\address[2]{Tomey Corporation, Nagoya 4510051, Japan}

\address[3]{Department of Electronic and Information Engineering, Changchun University, Changchun 130022, China}

\address[4]{School of Computer Science, University of Nottingham Ningbo China, Ningbo 315100, China}





\begin{abstract}
Precise lens structure segmentation is essential for the design of intraocular lenses (IOLs) in cataract surgery. Existing deep segmentation networks typically weight all pixels equally under cross-entropy (CE) loss, overlooking the fact that sub-regions of lens structures are inhomogeneous (e.g., some regions perform better than others) and poor segmentation calibration of boundary regions through the pixel aspect. Clinically, experts annotate different sub-regions of lens structures with varying confidence levels, considering factors such as sub-region proportions, ambiguous boundaries, and lens structure shapes. Motivated by this observation, we propose an Adaptive Confidence-Wise (ACW) loss to group each lens structure sub-region into different confidence sub-regions via a confidence threshold from the unique region aspect, aiming to exploit the potential of expert annotation confidence prior. Specifically, ACW clusters each target region into low-confidence and high-confidence groups and then applies a region-weighted loss to reweigh each confidence group. Moreover, this paper designs an adaptive confidence threshold optimization algorithm to adjust the confidence threshold of ACW dynamically. Additionally, to better quantify the miscalibration errors in boundary region segmentation, we propose a new metric, termed Boundary Expected Calibration Error (BECE). Extensive experiments on a clinical lens structure AS-OCT dataset and other multi-structure datasets demonstrate that our ACW significantly outperforms competitive segmentation loss methods across different deep segmentation networks (e.g., MedSAM). Notably, our method surpasses CE with \textbf{6.13\%} IoU gain, \textbf{4.33\%} DSC increase, and \textbf{4.79\%} BECE reduction in lens structure segmentation under U-Net. The code of this paper is available at \href{https://github.com/XiaoLing12138/Adaptive-Confidence-Wise-Loss}{https://github.com/XiaoLing12138/Adaptive-Confidence-Wise-Loss}.
\end{abstract}

\end{frontmatter}


\section{Introduction}
According to the World Report on Vision of World Health Organization (WHO) in 2019, approximately 65.2 million people are suffering from visual impairment and blindness due to cataract \cite{world2019world,ZHANG2022102499,zhang2022machine}. The most effective treatment for cataract patients is cataract surgery, which replaces the natural lens with an intraocular lens (IOL). Specifically, a significant procedure in the cataract surgery is the accurate IOLs design, which is determined by accurate lens structure segmentation. Currently, ophthalmologists usually segment lens structures rely on their clinical expertise and knowledge. However, this mode is subjective and error-prone. Therefore, it is essential to develop computer-aided techniques to segment lens structures accurately to improve the precision and efficiency of IOLs design.

\begin{figure}[t]
\centering\includegraphics[width=1.0\textwidth]{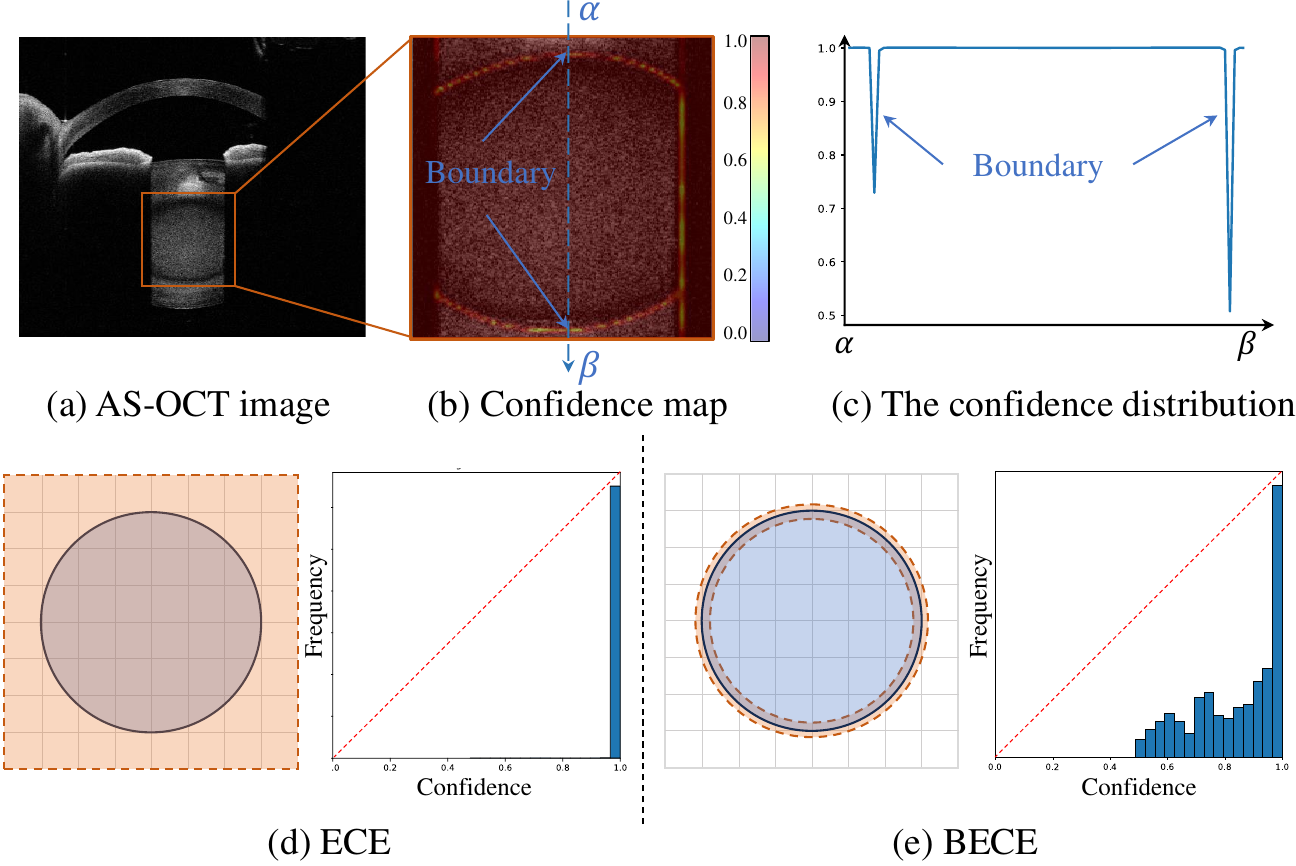}
\caption{The illustration of the confidence calibration problem in the lens structure segmentation task: (a) The AS-OCT image; (b) The confidence map near the nucleus region generated by U-Net; (c) The confidence distribution in the line $\alpha$ to $\beta$ where the confidence values of boundary region are smaller than confidence values of inner nucleus region; (d) The classical expected calibration error (ECE) is challenging to evaluate the segmentation calibration results well of boundary regions; (e) The boundary expected calibration error (BECE) focuses on confidence values of the boundary region, which can evaluate the segmentation calibration results near the boundary.}
\label{BECE}
\end{figure}

Recently, anterior segment optical coherence tomography (AS-OCT) has been an emerging ophthalmic imaging technique for cataract diagnosis and cataract surgery planning~\cite{zhang2024regional, wang2021objective, xiao2024multi}, which investigates the lens structures (e.g., nucleus, cortex, and capsule) clearly. With the advent of deep learning, deep segmentation networks have been applied to segment lens structures automatically based on AS-OCT images and have achieved good results. Yin et al.~\cite{yin2018automatic} employed the classical U-Net architecture for automatic lens structure segmentation. Zhang et al. \cite{zhang2019guided} proposed the Guided M-shape Convolutional Network (G-MNet) for automated lens structure segmentation, by introducing a guided block to emphasize low-contrast regions. \textit{However, these methods typically assign equal weights to all pixels in the lens structures based on standard cross-entropy (CE) loss, easily ignoring heterogeneous appearances of sub-regions in the lens structures from the region aspect. To be specific, they often achieve better segmentation results for the nucleus region compared to the capsule region in AS-OCT images.} To address this problem, some improved CE methods have been applied to improve the general segmentation results of different lens structures, such as weighted cross-entropy loss (WCE) \cite{ronneberger2015u}, focal loss (FL) \cite{lin2017focal}, distance map penalized cross-entropy loss (DPCE)\cite{caliva2019distance}. Unfortunately, current segmentation loss methods are incapable of accurately delineating pixels near the boundaries of different lens structures. As shown in Fig.~\ref{BECE}(b), the classical U-Net \cite{ronneberger2015u} exhibits higher uncertainty and poorer confidence calibration scores in the boundary regions of the nucleus, compared to the inner region. Surprisingly, we suggest that existing lens structure segmentation methods have overlooked this issue, motivating us to seek a new solution.


Moreover, experienced ophthalmologists often annotate different sub-regions of lens structures based on AS-OCT images with varying confidence levels in clinical practice. These confidence levels are affected by factors such as sub-region proportions, ambiguous or low-contrast boundaries, and the diverse shapes of lens structures. For example, ophthalmologists typically exhibit high confidence in accurately annotating the nucleus and cortex regions compared to the capsule region, where confidence levels are lower. This is primarily because the capsule region is smaller and more ambiguous than the nucleus and cortex regions. Consequently, to obtain high-quality annotated ground truth for all lens structures, ophthalmologists must devote extra attention to annotating the capsule region and ambiguous boundaries.

According to the above systematical analysis, one question naturally arises: \textit{Can we transform the expert annotation confidence prior into loss function design to guide deep segmentation networks in rebalancing the segmentation results of all lens structures and boosting the confidence levels in boundary region?}

Inspired by this problem, this paper introduces a novel Adaptive Confidence-Wise (ACW) loss, which divides pixel-wise prediction probability values in each segmented lens structure region into specific confidence sub-regions based on the adaptive confidence thresholds from the region aspect, by aiming to fully leverage the potential of expert annotation confidence prior. Specifically, ACW consists of two main steps: the clustering step and the weighting step. The clustering step clusters pixel-wise predicted probability values of each segmented lens structure into low-confidence and high-confidence groups via the adaptive confidence threshold. It is followed by the weighting step, a well-designed region-weighted loss is applied to obtain the loss of each confidence group, adjusting the loss proportions of each sub-lens structure region
accordingly. In particular, we designed an adaptive confidence threshold optimization (ACTO) algorithm to adjust the confidence threshold dynamically. This dynamic adjustment aims to further leverage the advantages of ACW in better rebalancing segmentation results of different lens structures, as well as improving confidence calibration results of boundary regions.


It is well-known that modern deep neural networks~\cite{guo2017calibration, zhong2021mislas,patra2023calibrating,wang2023calibrating,ye2023vsr}, including deep segmentation networks, are poorly calibrated in their predicted results. This is a significant security problem in computer-aided diagnosis systems. Several confidence calibration metrics have been proposed to measure the segmentation calibration performance of deep segmentation networks, e.g., Expected Calibration Error (ECE), but these evaluation metrics have limitations in evaluating the segmentation calibration performance of boundary regions well. Therefore, we introduce a new confidence calibration evaluation measure, dubbed Boundary Expected Calibration Error (BECE), to better measure the boundary region segmentation calibration performance, as shown in Fig.~\ref{BECE}(e).
In summary, our contributions are as follows: 
\begin{itemize}   
\item This paper proposes a novel Adaptive Confidence-wise (ACW) loss to improve both lens structure segmentation results and boundary region segmentation calibration performance by fully exploiting the potential of expert annotation confidence prior \textbf{from the region aspect}. 

\item We develop an ACTO algorithm to adjust the confidence threshold values in ACW adaptively. Additionally, we introduce the Boundary Expected Calibration Error (BECE) to better quantify the boundary region segmentation calibration performance of deep segmentation networks.

\item The comprehensive experiments on a clinical Lens Structure AS-OCT dataset and two other multi-structure datasets (Multi-structure AS-OCT and Synapse~\cite{landman2015miccai}) across three different deep segmentation networks, e.g., MedSAM~\cite{ma2024segment}, demonstrate that our ACW significantly outperforms other advanced segmentation losses in terms of segmentation and calibration performance. Further analysis explains the effectiveness of our method.
\end{itemize}

\section{Related Work}

\subsection{OCT Image-based Segmentation}
OCT imaging has become an indispensable tool for diagnosing ocular diseases, allowing for the detailed investigation of both the eye's posterior and anterior structures. With the advancement of deep learning, deep segmentation networks have been extensively to various eye structure segmentation based on OCT images, for better diagnosing ocular diseases, such as glaucoma and cataract \cite{ZHANG2022102499,hao2021hybrid,fu2019angle}. For posterior structure segmentation, Roy et al. \cite{roy2017relaynet} proposed a ReLayNet for the automated segmentation of retinal layers and fluids. Lee et al. \cite{lee2017deep} applied U-Net to segment the intraretinal fluid (IRF) structure. Wu et al.~\cite{wu2022choroidal} proposed a boundary enhancement network for automated choroidal layer segmentation. Yan et al.~\cite{yan2023automatic} applied a context-efficient adaptive network to segment the choroid layer and achieved good segmentation results. Regarding anterior structure segmentation, Mathai et al. \cite{mathai2019learning} presented a CorNet to segment the corneal structures in AS-OCT. Similarly, Sun et al.~\cite{sun2024oct} implemented corneal structure segmentation via a deep segmentation network. Yang et al. \cite{10256117} developed a hierarchical attention network to segment ciliary muscle automatically based on AS-OCT images.

More related to this paper is the lens structure segmentation in AS-OCT images. Cao et al.~\cite{cao2020efficient} utilized an efficient semantic segmentation network to segment different lens structures. Fang et al. \cite{fang2023lens} introduced shape constraints into U-Net to improve the segmentation results of lens structures. Xiao et al. \cite{xiao2024mmunetmixedmlparchitecture} proposed a mixed MLP architecture for automated lens structure segmentation. Although those deep segmentation networks have good segmentation results for lens structures, they weigh all pixels equally based on CE. This equal pixel weighting view overlooks the heterogeneous appearances of sub-regions in lens structures. As a result, they easily achieve better nucleus region segmentation results than the segmentation results of the capsule region. An efficient solution to this problem is designing a more specialized loss function by considering these variations in lens structures.

\subsection{Segmentation Loss Methods}
Segmentation loss design is an important yet hot research direction in image segmentation, including medical image segmentation. Its key goal is to guide deep segmentation networks to obtain expected segmentation results via parameter optimization. Classical CE \cite{ronneberger2015u} and Dice loss \cite{faustofully} are two most commonly used segmentation loss methods regarding their effectiveness and simplicity. However, they equally treat the pixels among without considering their differences among all segmentation regions, unavoidably getting poor segmentation results for small segmentation regions and ambiguous boundaries~\cite{petit2021u,ma2021loss,el2021high}. 
Focal Loss (FL)~\cite{lin2017focal} advances CE by assigning the large weights to hard-segmented pixels and small weights to easy-segmented pixels.
Tversky loss \cite{salehi2017tversky} extends the Dice loss by incorporating a hyper-parameter to rebalance the weights of false positive and false negative pixels. Following this, Abraham et al. \cite{abraham2019novel} combine the focal loss function with the Tversky index to produce the new FocalTversky loss. Moreover, Chen et al. \cite{chen2023adaptive} propose the Adaptive Region-Specific Loss (ARS), and Wang et al. introduce the Balancing Logit Variation Loss (BLV) to learn a more balanced representation across different categories. For lens segmentation, Fang et al. \cite{fang2023lens} proposed a level set loss to increase the weights of boundaries.

Additionally, we discovered that existing segmentation loss methods tend to be overconfident in boundary segmentation results, presenting a significant risk for subsequent disease diagnosis. This is mainly because these methods often neglect the different segmentation regions with varying confidence levels, particularly boundaries. Notably, regions with low confidence levels tend to be segmented poorly. Similarly, in clinical practice, experts often annotate each lens structure region with confidence levels and pay more attention to lens structure regions with lower confidence to ensure precise annotations. 
Inspired by this expert annotation confidence prior, we introduce an Adaptive Confidence-Wise (ACW) Loss, which is supposed to boost segmentation performance across different regions (particularly small segmentation regions) and confidence calibration of boundary regions without modifying deep segmentation networks from the region aspect rather than the pixel aspect, which has widely studied by current segmentation loss methods.


\section{Methodology}

\begin{figure}[t]
\centering\includegraphics[width=1.0\textwidth]{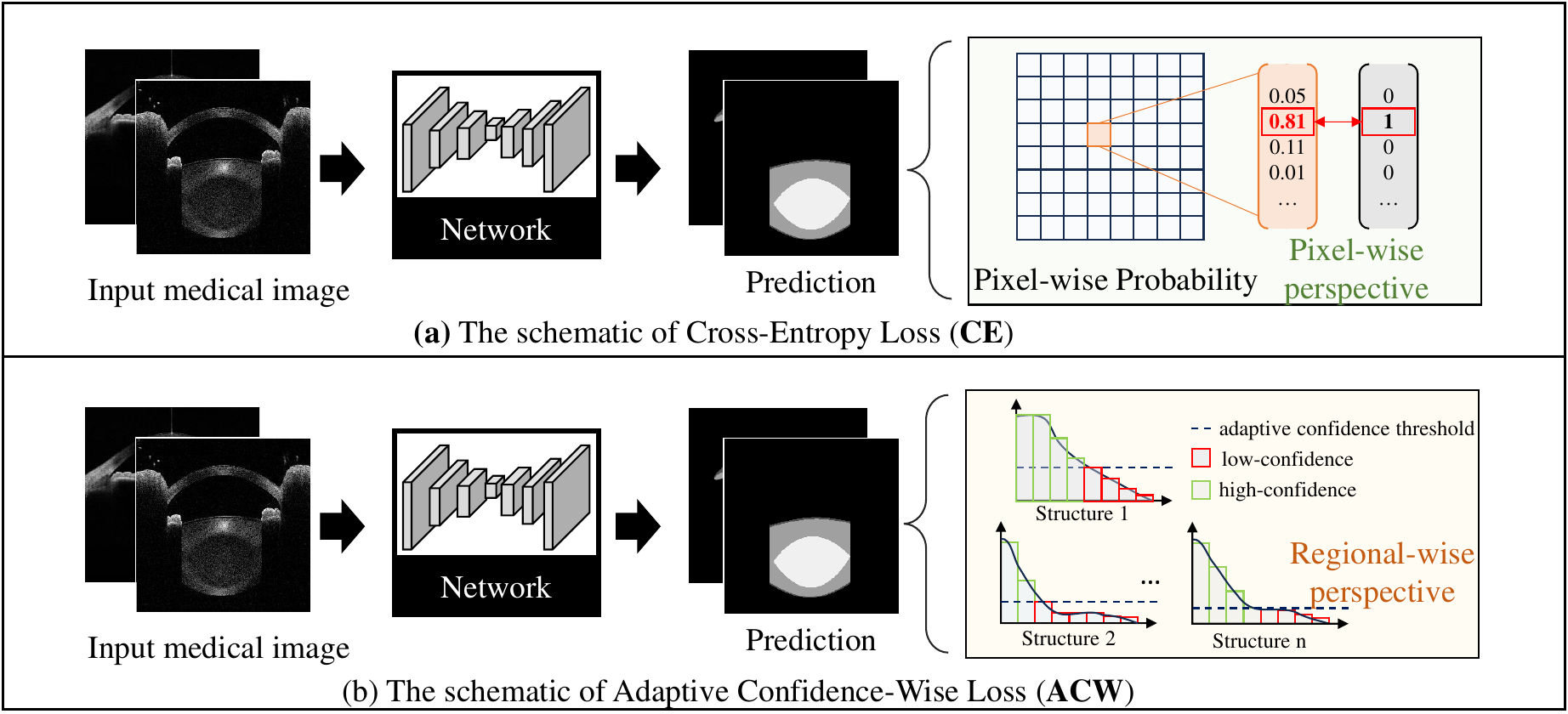}
\caption{Schematic comparison of Cross-Entropy Loss (CE) and Adaptive Confidence-wise Loss (ACW) in treating rediction probability values of all pixels in different segmentation regions: (1) CE assigns the same weights to prediction probability values of all pixels. (2) ACW first divides the prediction probability values of all pixels into low-confidence and high-confidence groups through the adaptive confidence threshold, then region-based weights is set to different confidence groups.} 
\label{ACW}
\end{figure}

\subsection{Rethinking Classical Cross-Entropy Loss}

Cross-Entropy (CE) loss is derived from Kullback-Leibler (KL) divergence, aiming to measure the dissimilarity between two probability distributions. In medical image segmentation, CE is calculated on a pixel-wise basis, representing the discrepancy between the predicted and ground truth distributions, as illustrated in Fig.~\ref{ACW}(a). Specifically, given an AS-OCT image $X\in \mathbf{R}^{C\times H\times W}$ as the input, the deep segmentation network (e.g., U-Net) produces the pixel-wise prediction probability values $P\in \mathbf{R}^{C\times H\times W}$, and the ground truth $Y\in \mathbf{R}^{C\times H\times W}$,  CE calculates as follows:
\begin{equation}
 L_{CE} = -\frac{1}{H\times W}\sum_{s=1}^{H \times W} \sum_{i=1}^{C} y_{s}^i\log (p_{s}^i), 
 \label{eq2} 
\end{equation}
where $H$ and $W$ denote AS-OCT image resolution, $p_s^i$ present the prediction probability of pixel $s$ belong to segmentation region $i$, and $y_{s}^i$ denote the ground truth. Taking $s-th$ pixel with the prediction probability value as the example, yielding the gradient value based on $L_{CE}$:
\begin{equation}
 \frac{\partial L_{CE(seg)}}{\partial p_{si}} = -\frac{1}{H \times W}\sum_{s=1}^{H \times W} \sum_{i=1}^{C} \frac{y_{si}}{p_{si}}.
 \label{eq3}
\end{equation}

According to Eq.~\ref{eq2} and~\ref{eq3}, CE commonly assigns equal weights to all pixels, overlooking the unique appearances and areas of different sub-regions in the lens structures. As a result, deep segmentation networks tend to segment the nucleus region more accurately than the capsule region, posing a significant challenge for subsequent clinical diagnosis and surgery planning. Although Several improved segmentation loss methods have been developed to alleviate this issue, they are still overconfident in segmenting boundaries and small lens structure regions, which is also a significant yet easily ignored issue in other multi-structure segmentation tasks.

\subsection{Adaptive Confidence-wise Loss}
Inspired by the expert annotation confidence prior, we propose a novel Adaptive Confidence-Wise loss (ACW) to improve the lens structure segmentation results and confidence calibration performance of boundary regions as illustrated in Fig.~\ref{ACW}(b). Unlike CE and its variants, ACW clusters the pixel-wise prediction probability values of each segmentation region into low-confidence and high-confidence groups with an adaptive confidence threshold optimization algorithm from the region aspect. Our ACW comprises two steps: clustering and weighting, which will be elaborated in detail.

\textbf{Clustering.} 
In the segmentation process, the pixel-wise prediction probability values generated by deep segmentation networks is not uniformly distributed, and they can be roughly classified into two groups: high-confidence and low-confidence. High-confidence group, typically denoting easily segmented regions, exhibit high segmentation accuracy with prediction confidence close to 1. Conversely, low-confidence group, characterized by poor segmentation results and low prediction confidence, which are often found in boundary regions.

Therefore, the ACW clusters pixel-wise prediction probability values of each segmented structure into two independent groups: high-confidence $R_{h}$ and low-confidence $R_{l}$, by setting a confidence threshold. Specifically, we set a confidence threshold to determine pixel-wise prediction probability values of different segmented structures belonging to a high-confidence group or a low-confidence group:
\begin{equation}
 R_{h}^i = \{s|p^i(s)>theshold^i\}, R_{l}^i = \{s|p^i(s)<threshold^i\},
 \label{eq3.1} 
\end{equation}
where $R_{h}^i$, $R_{l}^i$ and $threshold^i$ represent the high-confidence region, low-confidence region and $threshold^i$ of $i-th$ structure. 

\textbf{Weighting.} 
Given that professional ophthalmologists pay more attention to structure region with low confidence levels during the annotating process. Similarly, segmentation regions with low pixel-wise prediction probability value surely require greater attention. Motivated by this link, we develop a region-weighted loss to assign different weights to the high-confidence and low-confidence groups within each segmented structure region. The weighted loss enables deep segmentation networks to focus more on low-confidence regions during training, which is defined as:
\begin{equation}
L_{ACW}^i = -\frac{W_h}{|R_{h}^i|}\sum_{s \in R_{h}^i} y_s^i\log (p_s^i)-\frac{W_l}{|R_{l}^i|}\sum_{s \in R_{l}^i} y_s^i \log (p_s^i),
 \label{eq4}
\end{equation}
where \(|\cdot|\) represents the number of pixels, and the hyperparameter \(0 \leq \alpha \leq 1\) is used to control \(W_h\) and \(W_l\):

\begin{equation}
W_h = 1 - \alpha, W_l = 1 + \alpha,
 \label{eq5}
\end{equation}
Let \(\beta = \frac{|R^i_h|}{|R^i|}\) denote the ratio of the high-confidence region to the total region of class $i$. Under this definition, \(L^i_{ACW}\) can be reformulated as:

\begin{equation}
   \begin{aligned}
    &L_{ACW}^i = -\frac{1}{|R^i|}\big(\sum_{s \in R^i} y_s^i\log (p_s^i)\\
    &\underbrace{-\frac{\alpha+\beta-1}{\beta}\sum_{s\in R^i_h}y_s^i\log (p_s^i)}_{\text{Suppression Term}}\underbrace{+\frac{\alpha+\beta}{1-\beta}\sum_{s\in R^i_l}y_s^i\log (p_s^i)}_{\text{Enhancement Term}}\big),\\
    \end{aligned} 
\end{equation}
This reformulation highlights that \(L_{ACW}^i\) introduces a suppression term to downweight contributions from pixels in the high-confidence region while incorporating an enhancement term to emphasize contributions from pixels in the low-confidence region. These mechanisms collectively balance the optimization process by prioritizing regions that are more likely to exhibit uncertainty.
Specifically, We set \(\alpha = 0.4\) in our ACW, and more details about the settings of \(\alpha\) will be provided in the ablation study.
Finally, the ACW is calculated by the average of each $L_{ACW}^i$ as:

\begin{equation}
\begin{aligned}
& L_{ACW} = -\frac{1}{C}\sum_{i=1}^C\frac{1}{|R^i|}(T^i-\frac{\alpha+\beta-1}{\beta}T^i_h+\frac{\alpha+\beta}{1-\beta}T^i_l),\\
& where \quad T^i=\sum_{s\in R^i}y_s^ilog(p^i_s), 
\end{aligned}
 \label{eq5}
\end{equation}
compared to the CE loss \(L_{CE}\), which treats all regions equally under pixel aspect, \(L_{ACW}\) first aggregates the class-wise segmentation loss and weights each region by the corresponding class-wise pixel numbers. This manner helps mitigate the imbalance between large and small regions.

\begin{algorithm} 
    \caption{The adaptive confidence threshold optimization algorithm.} 
    \label{alg1} 
    \begin{algorithmic}
        \STATE \textbf{Initialization}: $R_{h}$, $R_{l}$, Data $D=\{x_1, \dots, x_n\}$, Network $f(,\theta)$
        \REPEAT
        \STATE $R_{h}=\emptyset$, $R_{l}=\emptyset$
        \STATE randomly choose $n$ samples from $D$ to construct $X$  
        \STATE $P \gets f(X,\theta)$
        \STATE $threshold = Q_{80}(P)$
        \FOR{$p_{s} \in P$}
        \IF{$p_{s} > threshold$} 
        \STATE $R_{h} \gets R_{h} \cup \{s\}$
        \ELSE 
        \STATE $R_{l} \gets R_{l} \cup \{s\}$
        \ENDIF
        \ENDFOR
        \STATE $L_{ACW}=-\frac{1}{C}\sum_{i=1}^C\frac{1}{|R^i|}(T^i-\frac{\alpha+\beta-1}{\beta}T^i_h+\frac{\alpha+\beta}{1-\beta}T^i_l)$
        \STATE update $\theta$
        \UNTIL $\theta$ is converge
    \end{algorithmic} 
\end{algorithm}

\textbf{Adaptive Region Clustering Scheme.} The pixel-wise prediction probability values in each segmentation region generated by deep segmentation networks fluctuate dynamically, significantly affecting the performance of our ACW. Consequently, an adaptive threshold strategy is necessary to differentiate between high-confidence and low-confidence groups accurately, by mocking re-check step in the expert annotation process. This threshold strategy plays a crucial role in rebalancing distributions of pixel-wise prediction probability values in segmentation regions. By dynamically adjusting to the varying confidence levels during training, the threshold strategy allows the deep segmentation networks to emphasize the boundary regions and small segmentation regions, ultimately improving segmentation performance and confidence calibration results of boundary regions.

Therefore, this paper proposes an Adaptive Confidence Threshold Optimization (ACTO) algorithm to achieve the above goal, as listed in Algorithm~\ref{alg1}. In the ACTO algorithm, we dynamically set the confidence threshold to $Q_{80}(P)$, where $Q_{80}(P)$ denotes the $80_{th}$ percentile of the predicted probability values. 
Therefore, the pixels number of $R_{high}$ and $R_{low}$ groups are expected to keep a fixed ratio of $8:2$, avoiding all pixels may fall into one group that causes a collapsing problem and harm the effectiveness of ACW during training. 

\textbf{Discussion:} Classical FL seems to be similar to ACW, which might mislead audiences, but we argue that they are different from each other. Here, we illustrate their main difference as follows: 
\begin{itemize}
   \item Our ACW improves the lens structure segmentation performance from the region aspect through comparisons to FL under the pixel aspect.
   \item ACW first introduces an ACTO algorithm to dynamically divide the pixel-wise prediction probability values of each segmentation region into low-confidence and high-confidence groups, then employs a region-weighted loss to assign different weights to these two confidence groups. That is, ACW assigns different weights to each confidence group, involving dynamic pixel-wise prediction probability value range. In contrast, FL assigns static weights to each pixel-wise prediction probability value.
\end{itemize}

\section{Experiment Settings and Metrics}

\subsection{Datasets and Implementation Details}
\subsubsection{Datasets}

To validate the effectiveness of our ACW, we adopt two private AS-OCT datasets and one public dataset for comparison. Specifically, the collection of these two private datasets complies with the tenets of the Helsinki Declaration.  

\textbf{Lens AS-OCT Dataset.} It is a private Lens AS-OCT dataset with 857 images from 115 eyes of 78 individuals. Among these, 532 AS-OCT images are from cataract patients, and 325 AS-OCT images are from healthy individuals. Each image is annotated with ground truth segmentations for all lens structures, including the capsule, cortical, and nucleus regions, as provided by three experienced ophthalmologists. We divided it into a training subset and a test subset in the ratio of $8 (62):2 (16)$ based on the individual level, ensuring no overlap between the subsets.

\textbf{Multi-structure AS-OCT Dataset.} It is another private dataset, comprising 288 AS-OCT images of healthy eyes, which is collected under darkroom conditions. These AS-OCT images are annotated with ground truth of the left iris, right iris, lens, and anterior chamber structures by three experienced clinicians. The dataset is also randomly partitioned into training and test subsets at an $8:2$ ratio based on the individual level.

\textbf{Synapse Dataset.} It is a public multi-organ dataset consists of 30 CT scans for multi-organ segmentation, which contains 8 abdominal organs (aorta, gallbladder, spleen, left kidney, right kidney, liver,
pancreas, spleen, stomach) as introduced in \cite{landman2015miccai}. 
The dataset is split into 18 training and 12 test samples by following the previous work~\cite{chen2021transunet,Zhang2023ppcr}.

\begin{table}[h]
\caption{Distribution of the dataset divisions. (Parentheses represent the number of individuals included).}
\label{Dataset}
\centering
    \begin{tabular}{c|c|c}
        \hline
           Dataset &  Train& Test\\
       \hline
            Lens AS-OCT & 657 (62)& 200 (16)\\
            Multi-structure AS-OCT & 252 (14)& 108 (6)\\
            Synapse & 1280 (18)& 847 (12)\\
        \hline        
    \end{tabular}
\end{table}

\subsubsection{Implementation Details}
Our proposed method and comparable methods are implemented by PyTorch and OpenCV packages. We train all methods on a 12G Titan V GPU and adopt the Adam optimizer to optimize the parameters of deep segmentation networks during training. We employ the Cosine Annealing Learning Rate strategy to adjust the learning rate, with an initial value set to 0.0015 \cite{loshchilov2016sgdr}.
The maximum number of training epochs is set to 100. During data preparation, we resize all images to 256$\times$256 pixels and use a random flip operation for data augmentation.

\subsection{Baselines}
To prove the effectiveness of our ACW comprehensively, we adopt competitive segmentation loss methods for comparison: CE, Dice Loss, Tversky Loss, FocalTversky~\cite{salehi2017tversky}, ARS \cite{chen2023adaptive}. These loss methods are evaluated by taking three representative deep segmentation networks as backbones: U-Net~\cite{ronneberger2015u}, TransUNet\cite{chen2021transunet} and MedSAM\cite{ma2024segment}.

\subsection{Evaluation Metrics}
This paper evaluates the general performance of ACW and comparable loss methods from two perspectives qualitatively and quantitatively: segmentation metrics and confidence calibration metrics.

\textbf{Segmentation Metrics}.
We adopt three commonly accepted evaluation metrics for measuring segmentation results: Dice score (DSC), IoU and HD95. Dice and IoU measure the overlap between the predicted segmentation results and the ground truth, and HD95 evaluates the spatial distance between the edges of the prediction and the ground truth.

\textbf{Confidence Calibration Metrics.}
Expected Calibration Error (ECE) is a classical metric proposed to assess confidence calibration performance in classification tasks. ECE is calculated as a weighted average of the absolute difference between accuracy (acc) and confidence (conf), given by:
\begin{equation}
    ECE = \sum_{m=1}^{M} \frac{\left| B_m \right|}{n}\left| acc(B_m) - conf(B_m) \right|,
\end{equation}
where the data is divided into \(M\) equally spaced bins. Moreover, \(B_m\) represents the \(m\)-th bin, \(\left|B_m\right|\) is the number of samples in that bin, and \(acc(B_m)\) and \(conf(B_m)\) denote the average accuracy and average confidence within bin \(B_m\), respectively.
Although ECE can be directly applied to segmentation tasks to evaluate overall segmentation calibration performance, it has limitations in assessing boundary region segmentation calibration, which is particularly crucial for auxiliary disease diagnosis.

Therefore, we propose a new confidence calibration metric, Boundary Expected Calibration Error (BECE), specifically designed to better assess the calibration performance of deep segmentation networks in boundary regions. Unlike ECE, BECE focuses solely on the confidence of pixels near the boundary, ignoring those within the inner region. The boundary region is defined as:

\begin{equation}
    M_{bd} = \bigcup_{c=1}^C Y_c \cap Dilation(1-Y_c), 
\end{equation}

\begin{equation}
    BECE = \sum_{B \in M_{bd}}\frac{\left| B_m \right|}{n}\left| acc(B_m)-conf(B_m)\right|, 
\end{equation}
where $Y_c$ denotes the label of class $c$, and $M_{bd}$ represents the boundary region, which is calculated by the union of the target region and the corresponding dilated background.


\section{Results and Discussion}

\subsection{Comparisons to State-of-the-art Segmentation Loss Methods}
Table~\ref{tab1} presents the lens structure segmentation results of three sub-regions of different segmentation loss methods on the Lens AS-OCT dataset. All segmentation loss methods adopt the same experiment settings to ensure a fair comparison. Our ACW consistently outperforms other state-of-the-art segmentation loss methods in terms of lens structure segmentation and confidence calibration performance across three deep segmentation network backbones: U-Net, TransUNet, and MedSAM. We also see: (1) For lens structure segmentation results, ACW obtains significant improvements with \textbf{6.13\%} in the IoU and \textbf{4.33\%} in the whole DSC, through comparisons to CE and Dice losses in U-Net. Based on the segmentation results of the capsule region, ACW achieves gains over \textbf{6.53\%}, \textbf{1.86\%} and \textbf{8.15\%} of IoU under U-Net, TranUNet, and MedSAM, demonstrating the superiority of our method in segmenting small lesion regions accurately. (2) For confidence calibration results, our ACW gets the lowest ECE and BECE compared to the other eight segmentation loss methods. Remarkably, compared to CE, ACW reduces BECE by over \textbf{4.79\%} in U-Net.
Overall, the results manifest the effectiveness of the ACW in improving lens structure segmentation and confidence calibration results.

In addition, we see that ACW with the U-Net performs better than it with TransUNet and MedSAM. Although MedSAM is a medical segmentation foundation model, it does not meet our expectations. One possible reason is that lens structure segmentation is specialized, containing small segmentation regions and low-contrast boundaries. Based on the results in Table~\ref{tab1}, we use U-Net as the backbone for the following ablation experiments.

\begin{table*}[h]
\caption{Performance comparisons of different loss methods on Lens AS-OCT in terms of segmentation metrics and confidence calibration metrics (Calib. Metrics) across three different backbones. (The red numbers mark the improvement compared to the second best result.)}
\label{tab1}
\centering
\scalebox{0.6}{
\begin{tabular}{c|c|cccccc|cc}
\hline
\multirow{3}{*}{Backbone} & \multirow{3}{*}{Loss} & \multicolumn{6}{c|}{Segmentation Metrics} & \multicolumn{2}{c}{Calib. Metrics} \\
\cline{3-10}
 & & \multicolumn{4}{c}{DSC $\uparrow$} & \multirow{2}{*}{IoU$\uparrow$}  & \multirow{2}{*}{HD95$\downarrow$}  & \multirow{2}{*}{ECE$\downarrow$}  & \multirow{2}{*}{BECE$\downarrow$}  \\
\cline{3-6}
 & & Nucleus & Cortex & Capsule & Whole & & & & \\
\hline
\multirow{7}{*}{U-Net}
& CE & 97.07 & 95.32 & 71.28 & 87.89 & 80.24 & 5.94 & 0.31 & 10.45 \\
& FL & 97.88 & 96.62 & 77.23 & 90.58 & 84.08 & 5.16 & 0.30 & 8.78 \\
& Dice & 97.75 & 96.14 & 50.77 & 81.55 & 74.06 & 39.43 & 0.92 & 27.07 \\
& ARS & 95.65 & 94.10 & 48.54 & 79.43 & 70.85 & 44.18 & 1.35 & 34.03 \\
& Tversky & \textbf{97.91} & 96.74 & 50.34 & 81.66 & 74.41 & 38.50 & 0.84 & 25.85 \\
& FocalTversky & 97.63 & 95.66 & 60.85 & 84.71 & 76.93 & 42.23 & 0.89 & 26.12 \\
& ACW & 97.87 & \textbf{96.83} & \textbf{81.96}\textcolor{red}{(\textbf{+4.73})} & \textbf{92.22}\textcolor{red}{(+1.64)} & \textbf{86.37}\textcolor{red}{(+2.29)} & \textbf{4.63} & \textbf{0.22} & \textbf{5.66}\textcolor{red}{(\textbf{-3.12})} \\
\hline
\multirow{7}{*}{TransUNet}
& CE & \textbf{97.87} & 96.74 & 80.05 & 91.55 & 85.41 & \textbf{5.29} & 0.25 & 7.04 \\
& FL & 97.76 & 96.56 & 79.18 & 91.17 & 84.83 & 5.53 & 2.21 & 6.60 \\
& Dice & 97.81 & 91.90 & 80.34 & 90.01 & 82.62 & 23.51 & 0.93 & 13.40 \\
& ARS & 97.62 & 96.39 & 80.83 & 91.61 & 85.40 & 6.45 & 0.51 & 16.21 \\
& Tversky & 97.28 & 94.85 & 80.46 & 90.86 & 84.07 & 8.29 & 0.48 & 11.89 \\
& FocalTversky & 97.70 & 94.59 & 79.94 & 90.74 & 83.94 & 14.36 & 0.78 & 16.62 \\
& ACW & 97.77 & \textbf{96.79} & \textbf{82.14}\textcolor{red}{(+1.31)} & \textbf{92.24}\textcolor{red}{(+0.63)} & \textbf{86.38}\textcolor{red}{(+0.97)} & 5.62 & \textbf{0.24} & \textbf{6.34}\textcolor{red}{(-0.26)} \\
\hline
\multirow{7}{*}{MedSAM}
& CE & 96.35 & 93.00 & 41.92 & 77.09 & 68.80 & 6.47 & 0.93 & 28.46 \\
& FL & 96.38 & 92.96 & 40.09 & 76.48 & 68.31 & 6.55 & 1.02 & 30.25 \\
& Dice & 96.32 & 93.09 & 40.89 & 76.77 & 68.56 & 6.68 & 0.92 & 28.26 \\
& ARS & 96.20 & 92.47 & 36.74 & 75.14 & 67.06 & 7.03 & 1.29 & 34.74 \\
& Tversky & 96.53 & 92.96 & 41.72 & 77.07 & 68.83 & 6.58 & \textbf{0.90} & 27.35 \\
& FocalTversky & 96.11 & 92.94 & 43.23 & 77.43 & 68.97 & \textbf{6.39} & 1.06 & 30.32 \\
& ACW & \textbf{96.75} & \textbf{93.85} & \textbf{52.64}\textcolor{red}{(\textbf{+9.41})} & \textbf{81.08}\textcolor{red}{(\textbf{+3.65})} & \textbf{72.61}\textcolor{red}{(\textbf{+3.64})} & 6.62 & 0.92 & \textbf{26.86}\textcolor{red}{(-0.49)} \\
\hline
\end{tabular}
}
\end{table*}

\subsection{Comparisons to Previous Lens Structure Segmentation Methods}
Table ~\ref{lens} offers the lens structure segmentation results of previous lens structure segmentation networks and our proposed ACW. It can be observed that ACW obtains promising performance with \textbf{86.37\%} of IoU, \textbf{92.22\%} of DSC, and \textbf{4.63} of HD95 accordingly. For example, ACW outperforms \cite{cao2020efficient} by \textbf{19.83\%} of IoU, \textbf{15.36\%} of DSC, while decreasing over \textbf{0.6} of HD95 and \textbf{5.88\%} of BECE compared with others.
The results prove our ACW's effectiveness in boosting lens structure segmentation performance and reducing boundary errors via confidence sub-region clustering and confidence group weighting.

\begin{table}
\caption{Comparison of lens structure segmentation methods in terms of segmentation and confidence calibration metrics (Calib. Metrics).}
\label{lens}
\centering
\scalebox{0.75}{
\begin{tabular}{c|ccc|c}
\hline
\multirow{2}{*}{Method} & \multicolumn{3}{c|}{Segmentation Metrics} & \multicolumn{1}{c}{Calib. Metrics} \\
\cline{2-5}
 & IoU$\uparrow$ & DSC$\uparrow$ & HD95$\downarrow$ & BECE$\downarrow$ \\
\hline
\cite{zhang2019guided} & 84.00 & 90.50 & 5.23 & 11.54 \\
\cite{cao2020efficient} & 66.54 & 76.86 & 11.24 & 35.57 \\
\cite{fang2023lens} & 76.58 & 85.17 & 6.55 & 17.23 \\
ACW & \textbf{86.37} & \textbf{92.22} & \textbf{4.63} & \textbf{5.66} \\
\hline
\end{tabular}
}
\end{table}

\subsection{Generalization Validation}
To further to prove the generalization ability of our ACW, we conduct extensive experiments on two other multi-structure segmentation datasets. Here, we also adopt three different backbones.


\textbf{Results on Multi-structure AS-OCT dataset:} Table~\ref{MS} presents the multi anterior structure segmentation results on the multi-structure AS-OCT dataset in terms of IoU, Dice, HD95, and BECE. Our ACW consistently achieves the best results in both segmentation metrics and calibration metrics. Notably, ACW achieves a BECE improvement of over \textbf{8.07\%} through comparisons to Dice, Tversky, and FocalTversky across all three backbones. Additionally, compared to Focal, ACW obtains an IoU increase of over \textbf{0.44\%} across all three backbones. Additionally, based on the segmentation results of the Iris\_L and Iris\_R regions, which are smaller than other regions, ACW achieves DSC gains of 1.36\% and 1.54\% under U-Net, respectively, further demonstrating its effectiveness in segmenting small lesion areas accurately.

\begin{table}[ht]
\caption{Performance comparisons of different loss methods on Multi-structure AS-OCT in terms of segmentation and confidence calibration metrics (Calib. Metrics) under different backbones.}
\label{MS}
\centering
\scalebox{0.6}{
\begin{tabular}{c|c|ccccccc|c}
\hline
\multirow{3}{*}{Backbone} & \multirow{3}{*}{Loss} & \multicolumn{7}{c|}{Segmentation Metrics} & Calib. Metrics \\
\cline{3-10}
 & & \multicolumn{5}{c}{DSC $\uparrow$} & \multirow{2}{*}{IoU$\uparrow$}  & \multirow{2}{*}{HD95$\downarrow$}& \multirow{2}{*}{BECE$\downarrow$}  \\
\cline{3-7}
 & & AS & Iris\_L & Iris\_R & Lens &Whole & & & \\
\hline
\multirow{6}{*}{U-Net} 
 & CE           &99.10&94.02&93.91&98.66& 96.42 & 93.20 & 4.72  & 2.43 \\
 & Dice         &98.89&93.67&93.77&98.32& 96.16 & 92.72 & 6.65  & 14.52 \\
 & Focal        &\textbf{99.20}&93.90&94.06&98.77& 96.48& 93.32& 4.44& 4.91 \\
 & Tversky      &99.02&91.76&92.10&98.35& 95.31 & 91.23 & 7.82  & 14.42 \\
 & FocalTversky &98.60&89.34&89.49&97.80& 93.81 & 88.66 & 11.06 & 19.90 \\
& Ours & 99.10& \textbf{95.38} & \textbf{95.60} & \textbf{98.81} & \textbf{97.22} & \textbf{94.65} & \textbf{3.27} & \textbf{1.59} \\

\hline
\multirow{6}{*}{TransUNet} 
 & CE           &99.19&92.18&92.46&98.91& 95.68 & 91.92 & 5.45  & 3.08 \\
 & Dice         &99.13&95.65&95.66&98.50& 97.23 & 94.66 & 2.19  & 10.78 \\
 & Focal        &98.76&92.61&93.78&98.44& 95.90 & 92.25 & 7.13  & 10.39 \\
 & Tversky      &98.80&95.15&95.31&98.40& 96.91 & 94.07 & 3.16  & 13.38 \\
 & FocalTversky &99.03&95.29&95.33&98.20& 96.96 & 94.15 & 2.67  & 12.13 \\
 &Ours & \textbf{99.25} & \textbf{96.11} & \textbf{96.35} & \textbf{98.98} & \textbf{97.67} & \textbf{95.49} & \textbf{1.67} & \textbf{2.32} \\

\hline
\multirow{6}{*}{MedSAM} 
 & CE           &97.66&87.60&87.84&98.20& 92.82& 87.04& 6.25& 16.54\\
 & Dice         &97.30&87.07&87.43&97.40& 92.30 & 86.11 & 6.85  & 26.55 \\
 & Focal        &\textbf{97.70}&88.27&87.96&98.11& 93.01 & 87.32 & 5.79  & 11.14 \\
 & Tversky      &97.21&87.64&87.94&97.39& 92.55 & 86.49 & \textbf{5.45}& 27.82 \\
 & FocalTversky &96.97&86.73&86.40&97.66& 91.94 & 85.54 & 6.83  & 28.76 \\
 & Ours & 97.67& \textbf{88.41} & \textbf{88.74} & \textbf{98.28} & \textbf{93.27} & \textbf{87.76} & 5.69& \textbf{11.01} \\

\hline
\end{tabular}
}
\end{table}

\begin{figure*}[t]
\centering
\includegraphics[width=1.0\textwidth]{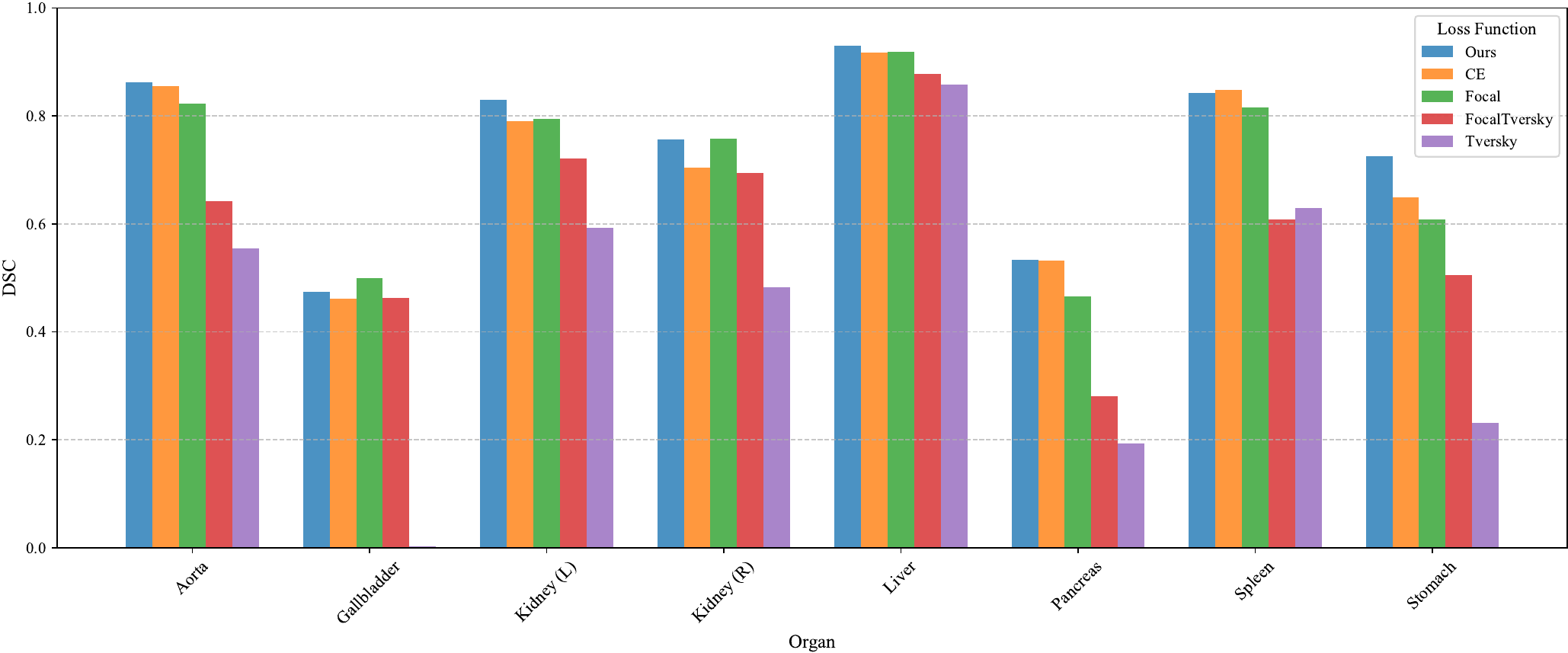}
\caption{Performance visulization of ACW and state-of-the-art loss methods on Synapse across different organs.} \label{syn}
\end{figure*}

\textbf{Results on Synapse dataset:} We present the multi-organ segmentation results of our method and other competitive loss functions in Table~\ref{synapse}. Remarkably, ACW achieves improvements of \textbf{3.08\%} in IoU and \textbf{2.48\%} in DSC, while reducing BECE by \textbf{3.28\%}, compared to competitive calibration loss functions across all three backbones. In particular, ACW demonstrates the best performance under the U-Net backbone, with a DSC of \textbf{81.62\%} and an IoU of \textbf{70.57}, achieving a reduction of \textbf{4.81} in HD95 compared to other methods. Furthermore, Fig.~\ref{syn} illustrates the DSC scores across all eight organs using the U-Net backbone. Notably, ACW achieves the highest DSC on the smallest organ, the Aorta, and also performs competitively on the largest organ, the Stomach. Overall, these results confirm that our method effectively enhances both segmentation accuracy and confidence calibration, aligning with our expectations.

\begin{table}
\caption{Performance comparisons of ACW and state-of-the-art loss methods on Synapse dataset in terms of segmentation and confidence calibration metrics (Calib. Metrics) under different backbones.}
\label{synapse}
\centering
\scalebox{0.9}{
\begin{tabular}{c|c|c|c|c|c}
\hline
\multirow{2}{*}{Backbone} & \multirow{2}{*}{Loss} & \multicolumn{3}{c|}{Segmentation Metrics} & Calib. Metrics\\
\cline{3-6}
 & & IoU$\uparrow$ & DSC$\uparrow$ & HD95$\downarrow$ & BECE$\downarrow$ \\
\hline
\multirow{5}{*}{U-Net}& CE&58.37&  71.94& 22.05& 15.53\\
 & Dice&21.04&  37.79 &57.35 & 49.28\\
 & Focal& 57.24&  71.03& 22.25& 16.83\\
 & Tversky& 32.15&  44.32& 54.18& 38.14\\
 & FocalTversky& 44.89&  59.92& 23.17& 37.52\\
 & Ours&\textbf{61.45}& \textbf{74.42} &\textbf{17.24} & \textbf{9.38}\\
\hline
\multirow{6}{*}{TransUNet} &CE&63.72 &72.18 &39.79 & 23.04\\
 & Dice&41.39& 51.94& 22.94& 29.07\\
 & Focal& 66.94&  77.92& 10.51& 12.32\\
 & Tversky&60.76& 70.96& 11.72& 22.08\\
 & FocalTversky& 61.72& 71.52& 10.34& 21.94\\
 & Ours&\textbf{70.59}&\textbf{81.62} &\textbf{9.48} & \textbf{9.04}\\
\hline
\multirow{6}{*}{MedSAM}&CE& 36.82& 49.85& 28.60& 39.73\\
 & Dice& 26.79& 34.20& 24.90& 44.34\\
 & Focal& 31.59& 44.69&26.76 & 32.52\\
 & Tversky& 16.09& 23.87& 46.75& 54.56\\
 & FocalTversky& 25.20& 35.08& 28.02& 49.93\\
 & Ours&\textbf{40.62} & \textbf{53.63} &\textbf{27.49} & \textbf{25.75}\\
\hline
\end{tabular}
}
\end{table}

\subsection{Ablation Study}
In this section, we conduct a series of experiments to investigate the impacts on the ACW.

\subsubsection{Effects of Rebalanced Weight $\alpha$} 
To investigate the influence of rebalanced weight $\alpha$, we test five different rebalanced weight values using the U-Net as the backbone, as listed in Table~\ref{weight}. We see that it is challenging to set the proper $\alpha$ value, when $\alpha$ is assigned to 0.4, ACW generally performs better than it with other $\alpha$ settings. Additionally, we find that assigning large weights to low-confidence regions that achieves better segmentation results and confidence calibration performance of boundary regions than those using equal region weights ($\alpha=0$).

\begin{table}
\caption{Results of different rebalanced weight $\alpha$ settings of ACW under U-Net in terms of segmentation and confidence calibration metrics (Calib. Metrics).}
\label{weight}
\centering
\begin{tabular}{c|ccc|c}
\hline
\multirow{2}{*}{$\alpha$} & \multicolumn{3}{c|}{Segmentation Metrics} & \multicolumn{1}{c}{Calib. Metrics} \\
\cline{2-5}
 & IoU $\uparrow$ & DSC $\uparrow$ & HD95 $\downarrow$ & BECE $\downarrow$ \\
\hline
0.0 & 80.24 & 87.89 & 5.94 & 10.45 \\
0.2 & 84.83 & 91.21 & 10.39 & 10.13 \\
0.4 & 86.37 & 92.22 & \textbf{4.63} & \textbf{5.66} \\
0.6 & \textbf{86.80} & \textbf{92.49} & 5.31 & 10.19 \\
0.8 & 86.52 & 92.29 & 5.15 & 12.64 \\
\hline
\end{tabular}
\end{table}

\subsubsection{Effects of Confidence-Level Threshold Selection} 
Table~\ref{threshold} presents lens segmentation result comparisons under different confidence level threshold selections of ACW. ACW achieves the best performance when the confidence level threshold is assigned to $Q_{80}$, yielding  \textbf{86.37\%} of DSC and \textbf{5.66\%} of BECE. The results verify that low-confidence regions typically contain a small proportion of pixels, which is consistent with our hypothesis.

\begin{table}
\caption{Results of different confidence level threshold selections of ACW under U-Net in terms of segmentation and confidence calibration metrics (Calib. Metrics).}
\label{threshold}
\centering
\begin{tabular}{c|ccc|c}
\hline
\multirow{2}{*}{Threshold} & \multicolumn{3}{c|}{Segmentation Metrics} & \multicolumn{1}{c}{Calib. Metrics} \\
\cline{2-5}
 & IoU$\uparrow$ & DSC$\uparrow$ & HD95$\downarrow$ & BECE$\downarrow$ \\
\hline
$Q_{10}$ & 84.05 & 90.69 & 5.14 & 9.34 \\
$Q_{20}$ & 83.37 & 90.13 & 5.66 & 10.76 \\
$Q_{30}$ & 85.73 & 91.83 & 4.88 & 7.08 \\
$Q_{40}$ & 85.69 & 91.81 & 4.76 & 8.87 \\
$Q_{50}$ & 85.34 & 91.49 & 6.28 & 8.89 \\
$Q_{60}$ & 85.78 & 91.82 & 5.58 & 8.95 \\
$Q_{70}$ & 86.00 & 91.96 & 5.41 & 10.35 \\
$Q_{80}$ & 86.37 & 92.22 & \textbf{4.63} & \textbf{5.66} \\
$Q_{90}$ & \textbf{86.88} & \textbf{92.54} & 5.60 & 12.50 \\
\hline
\end{tabular}
\end{table}

\subsubsection{Combination with Different Loss Methods} 
Table~\ref{combinations} presents the lens structure segmentation results of combining our method with various loss functions. Although different loss combinations improve IoU, they increase BECE. For example, the combination of ACW and Dice loss function achieves gains of \textbf{1.17\%} of IoU, accompanied by a \textbf{9.19\%} increase in BECE. One key reason to account for the results is that the combination loss methods may make deep segmentation networks overconfident.

\begin{table}[H]
\caption{Lens structure segmentation results of combining different segmentation loss functions based on U-Net, reported in terms of segmentation and confidence calibration metrics (Calib. Metrics).}
\label{combinations}
\centering
\begin{tabular}{c|ccc|c}
\hline
\multirow{2}{*}{Loss} & \multicolumn{3}{c|}{Segmentation Metrics} & Calib. Metrics \\
\cline{2-5}
 & IoU$\uparrow$ & DSC$\uparrow$ & HD95$\downarrow$ & BECE$\downarrow$ \\
\hline
ACW & 86.37 & 92.22 & \textbf{4.63} & \textbf{5.66} \\
+Focal & 86.74 & 92.43 & 5.00 & 11.48 \\
+Dice & \textbf{87.54} & \textbf{92.98} & 4.96 & 14.85 \\
+Tversky & 86.46 & 92.29 & 5.32 & 13.24 \\
+FocalTversky & 87.30 & 92.82 & 5.20 & 14.30 \\
\hline
\end{tabular}
\end{table}

\begin{figure*}[t]
\centering
\includegraphics[width=1.0\textwidth]{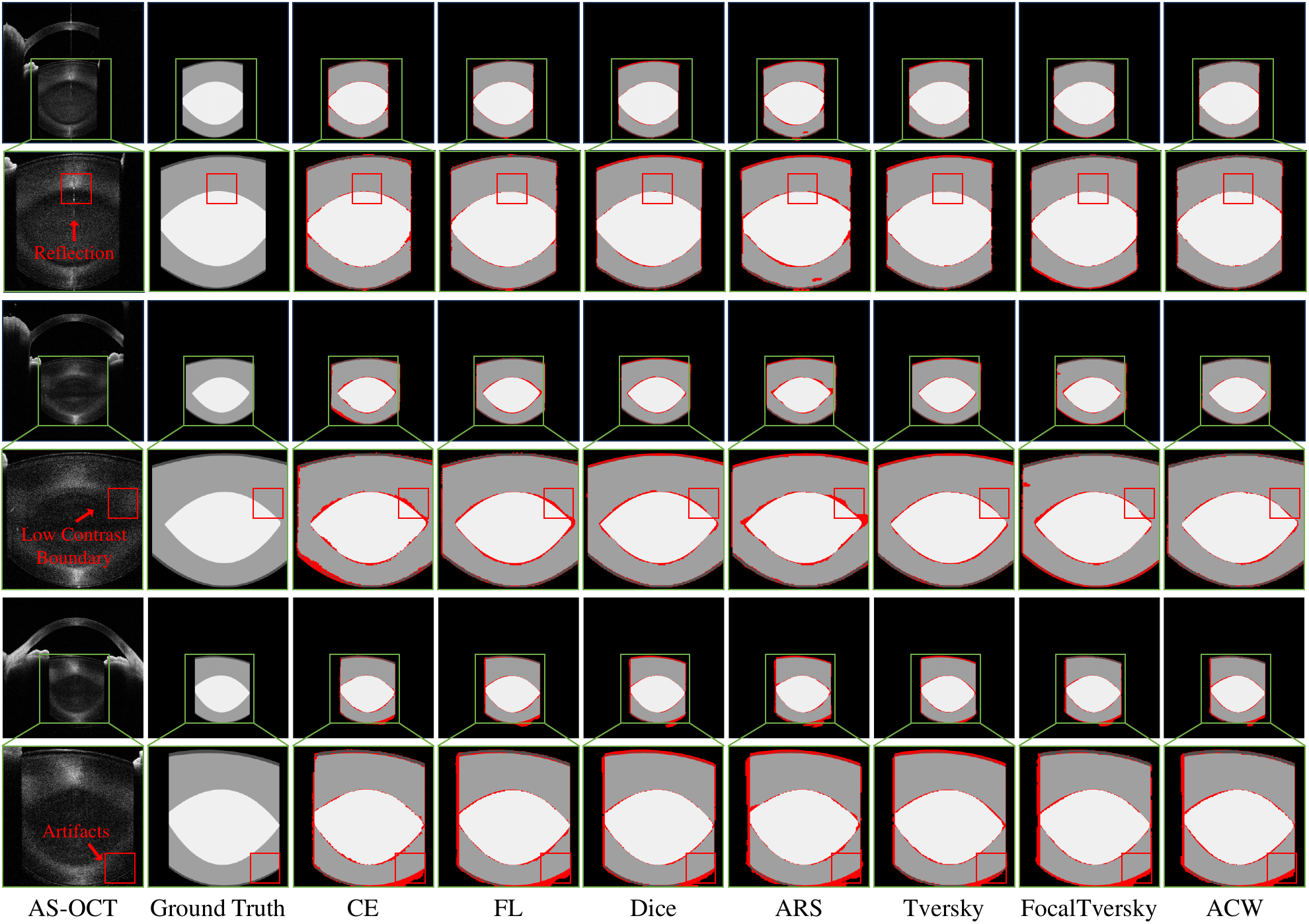}
\caption{Visual lens structure segmentation comparisons of our ACW and six other representative loss methods based on U-Net. Here, we take three representative AS-OCT images of lens structure with three sub-regions: reflection, low contrast boundary and artifacts. The red region indicates the incorrectly segmented pixels. Our ACW achieves promising lens segmentation results on the first two AS-OCT images ( Row 1 to Row 4), while it generates wrong segmentation results at the artifacts of the last AS-OCT image (Row 5 and Row 6). Generally, ACW performs better than other SOTA loss methods on three representative AS-OCT images.} \label{lens_well}
\end{figure*}

\subsection{Visualization Analysis and Explanation}

\subsubsection{Visualization Analysis of Lens Structure Segmentation}
Fig.~\ref{lens_well} presents a comprehensive analysis of lens structure segmentation visualizations among our ACW and six other representative loss methods based on the U-Net. In this paper, three representative AS-OCT images from the test set are selected, containing reflection (Row 1 and Row 2), low contrast boundary (Row 3 and Row 4) and artifacts (Row 5 and Row 6). Column 1 and Column 2 offers the original AS-OCT images and ground truth of three sub-lens structure annotations. Column 3 to Column 6 indicate the segmented lens structure results generated by our method and comparable loss methods. Our observations are summarized as follows: (1) The presence of reflections in the AS-OCT image has little negative effects on lens structure segmentation performance, verified by all loss methods under U-Net. (2) ACW demonstrates superiority in lens structure segmentation performance compared to other loss methods, reflected in two important areas: it performs better in segmenting capsule regions and significantly improves segmentation results near low-contrast boundaries. (3) Although ACW may lead to mis-segmentation in the presence of artifacts, it still outperforms better than other competitive methods.
The visual segmentation analysis proves that our ACW effectively improves different lens structure segmentation performance, especially small capsule regions by exploring expert annotation confidence prior.

\begin{figure*}[t]
\centering
\includegraphics[width=1.0\textwidth]{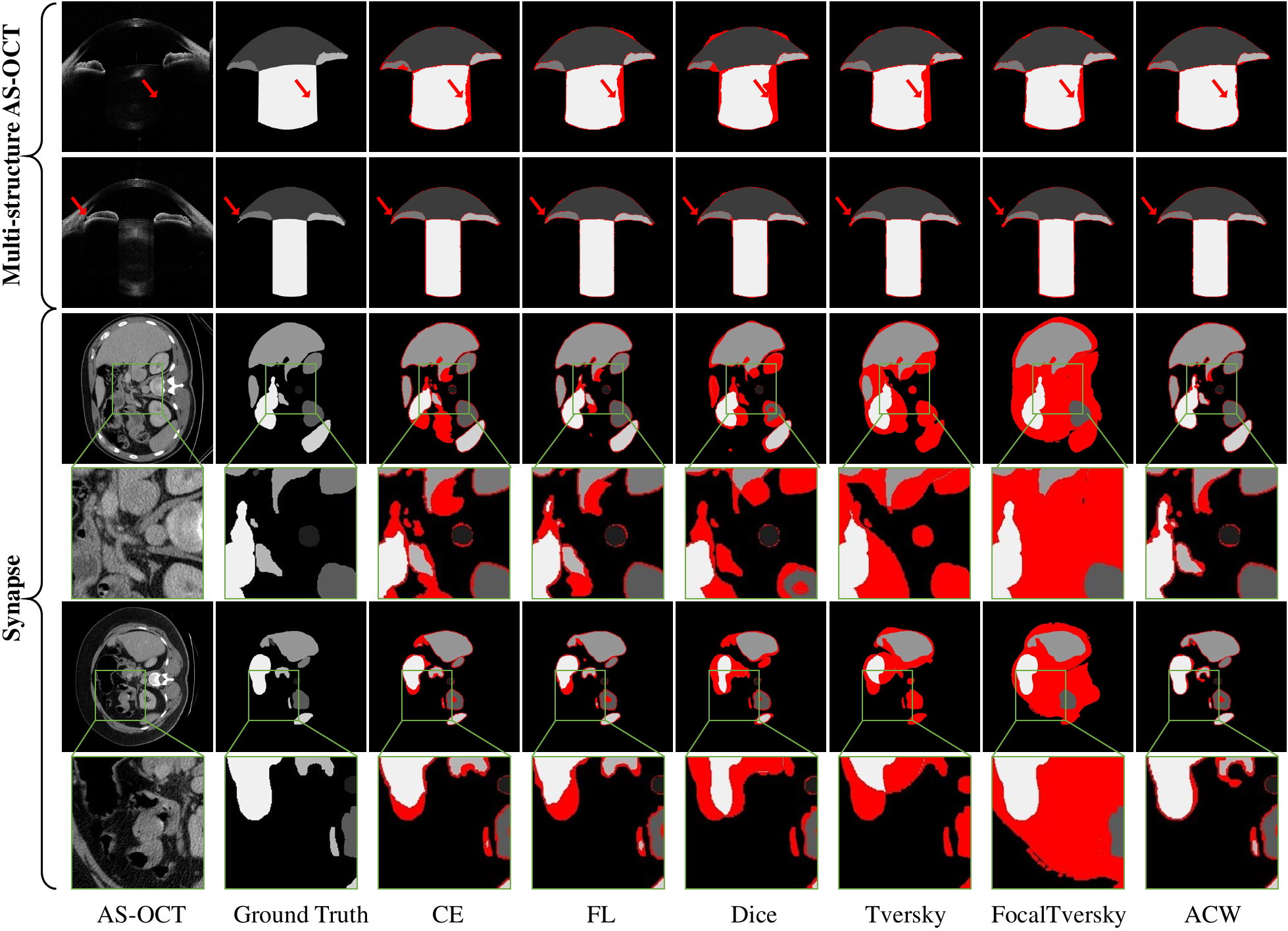}
\caption{Visual Multi-structure AS-OCT and Synapse dataset segmentation using our ACW method and six other representative loss methods based on
U-Net. We present two representative images for Multi-structure AS-OCT (Rows 1 to 2) and the Synapse dataset (Rows 3 to 6), respectively. The red
regions indicate incorrectly segmented pixels. In both datasets, ACW performs better than other state-of-the-art loss methods.} \label{other_res}
\end{figure*}

\subsection{Visualization Analysis of Multi-structure AS-OCT Dataset and Synapse Dataset}
Fig.~\ref{other_res} offers the visual segmentation comparisons of ACW and other SOTA loss methods on the Multi-structure AS-OCT Dataset and the Synapse Dataset. For the multi-anterior structure segmentation of the eye, our ACW significantly outperforms other methods, particularly around boundaries and small anterior regions such as the iris. Similarly, ACW achieves promising multi-organs results in segmenting small organs, whereas some other loss functions fail to segment these small segmentation regions accurately. The visual results further demonstrate the generalization ability of our method.

\subsubsection{Visualization Analysis of Boundary Region Confidence Calibration Results on Lens AS-OCT Dataset}
\begin{figure*}[t]
\centering
\includegraphics[width=1.0\textwidth]{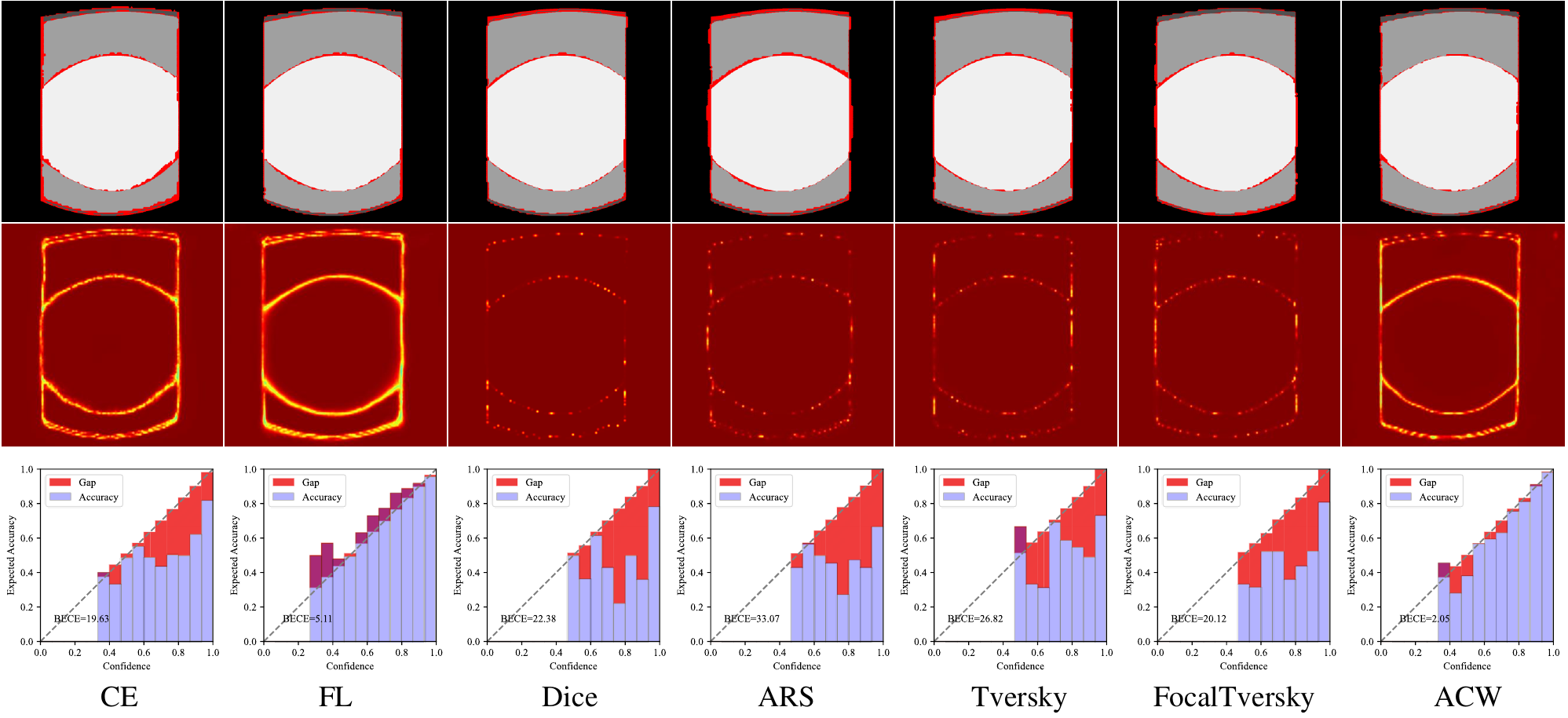}
\caption{Visualization of the calibration results of lens segmentation results within different losses. (The first row is the segmentation results, the second row is the confidence map and the third row is the reliability diagrams of BECE)} \label{cali}
\end{figure*}

We also visualize the segmentation calibration results and BECE between our ACW method and six other methods in Fig.~\ref{cali} via heat maps and reliability diagrams. The first row offers the lens structure segmentations produced by our ACW and other loss methods; the second row and the third row present confidence maps and reliability diagrams. We observe as follows: (1) Low-confidence regions are predominantly located around boundaries and are highly correlated with mis-segmentation regions, consistent with our expectations. (2) Confidence maps and reliability diagrams for BECE indicate that our ACW achieves better boundary region segmentation calibration results than other SOTA methods. According to Fig.~\ref{lens_well} - Fig.\ref{cali}, the visual results demonstrate that our method not only improves segmentation performance but also boosts segmentation calibration performance of boundary regions via clustering and weighting steps.

\section{Conclusion}
Motivated by expert annotation confidence prior, this paper proposes an Adaptive Confidence-Wise (ACW) loss to improve the lens structure segmentation results and boundary segmentation calibration performance from the unique region aspect. Additionally, the adaptive confidence threshold optimization (ACTO) algorithm is developed to better cluster pixel-wise predicted probability of each segmented sub-lens structure into low-confidence and high-confidence groups. The extensive experiments on the Lens AS-OCT dataset demonstrate that our method significantly outperforms four competitive segmentation loss methods across three backbones in terms of segmentation metrics and confidence calibration metrics. In the future, we plan to extend our method to unsupervised medical segmentation tasks and provide more theoretical analysis.

\section*{Acknowledgements}
This work was supported in part by National Key R\&D Program of China (No.2024 YFE0198100), National Key R\&D Program of China (No.2024YFC2510800), General Program of National Natural Science Foundation of China (Grant No.82272086) and Shenzhen Medical Research Fund (Grant No. D2402014).

\section*{Declaration of Competing Interest}
The authors declare that they have no competing financial interests or personal relationships that could have appeared to influence the work reported in this paper

\bibliography{mybibliography}

\end{document}